%% file: main.tex
\documentclass{article}

\usepackage[margin=1in]{geometry}

\usepackage[T1]{fontenc}
\usepackage[utf8]{inputenc}

\DeclareUnicodeCharacter{2212}{\ensuremath{-}}     
\DeclareUnicodeCharacter{2013}{--}                 
\DeclareUnicodeCharacter{2014}{---}                
\DeclareUnicodeCharacter{00D7}{\ensuremath{\times}}
\DeclareUnicodeCharacter{2009}{\,}                 

\usepackage{titlesec}
\usepackage{tocbibind}
\usepackage{enumitem}
\usepackage{booktabs}
\usepackage{graphicx}
\usepackage{tabularx}
\usepackage{longtable}  
\usepackage{array}
\usepackage{ragged2e}
\usepackage{ltablex}
\keepXColumns  

\usepackage{cite}

\usepackage{hyperref}
\hypersetup{colorlinks=true, linkcolor=blue, citecolor=blue, urlcolor=blue}

\makeatletter
\renewcommand{\maketitle}{%
  \begin{center}
    {\LARGE \@title \par}
    \vspace{1em}
  \end{center}
}
\makeatother

\newcommand{\totaldatasets}{54}
\newcommand{\totalsubjects}{538{,}031}
\newcommand{\selecteddatasets}{15}

\title{A Structured Review and Quantitative Profiling of Public Brain MRI
Datasets for Foundation Model Development}

\begin{document}

\maketitle

\input{sections/authors}
\input{sections/abstract}
\input{sections/introduction}
\input{sections/methodology}
\input{sections/dataset_level}
\input{sections/image_level}
\input{sections/preprocessing}
\input{sections/covariate_shift}
\input{sections/conclusion}
\input{sections/acknowledgement}

\clearpage
\bibliographystyle{ieee}
\bibliography{references}

\newpage
\input{sections/appendix}

\end{document}

%% file: sections/authors.tex
\begin{center}
    Minh Sao Khue Luu${}^{1,*}$, Margaret V. Benedichuk${}^{1}$, Ekaterina I. Roppert${}^{1}$, Roman M. Kenzhin${}^{1}$, Bair N. Tuchinov${}^{1}$\\
    \vspace{0.5em}
    ${}^1$
The Artificial Intelligence Research Center of Novosibirsk State University, 630090 Novosibirsk, Russia\\

    \vspace{0.5em}
    ${}^*$\texttt{khue.luu@g.nsu.ru}\\
\end{center}

%% file: sections/abstract.tex




\begin{abstract}

The development of foundation models for brain MRI depends critically on the scale, diversity, and consistency of available data, yet systematic assessments of these factors remain scarce. In this study, we analyze \totaldatasets{} publicly accessible brain MRI datasets encompassing over \totalsubjects{} to provide a structured, multi-level overview tailored to foundation model development. At the dataset level, we characterize modality composition, disease coverage, and dataset scale, revealing strong imbalances between large healthy cohorts and smaller clinical populations. At the image level, we quantify voxel spacing, orientation, and intensity distributions across \selecteddatasets{} representative datasets, demonstrating substantial heterogeneity that can influence representation learning. We then perform a quantitative evaluation of preprocessing variability, examining how intensity normalization, bias field correction, skull stripping, spatial registration, and interpolation alter voxel statistics and geometry. While these steps improve within-dataset consistency, residual differences persist between datasets. Finally, feature-space case study using a 3D DenseNet121 shows measurable residual covariate shift after standardized preprocessing, confirming that harmonization alone cannot eliminate inter-dataset bias. Together, these analyses provide a unified characterization of variability in public brain MRI resources and emphasize the need for preprocessing-aware and domain-adaptive strategies in the design of generalizable brain MRI foundation models.


\end{abstract}

%% file: sections/introduction.tex
\section{Introduction}

Brain diseases such as tumors, Alzheimer's disease, multiple sclerosis, and stroke affect millions worldwide, leading to significant health and societal burdens \cite{feigin_global_2019}. Magnetic resonance imaging (MRI) has become the central modality for studying these conditions due to its non-invasive nature, superior soft tissue contrast, and ability to capture diverse anatomical and physiological information across multiple sequences. While originally developed for clinical decision-making, the rapid expansion of publicly available MRI datasets has transformed neuroimaging into a data-driven domain where large-scale machine learning, particularly foundation models, now plays a pivotal role.

Foundation models, first established in natural language processing and computer vision \cite{bommasani_opportunities_2021}, are increasingly being explored for medical imaging \cite{azad_foundational_2023, he_foundation_2024, ma_segment_2024, linguraru_foundation_2024, cox_brainsegfounder_2024}. Their promise lies in learning generalizable representations from heterogeneous data and transferring them to a wide range of downstream tasks. However, the success of such models in brain MRI crucially depends on the availability of harmonized, large-scale datasets. Unlike other imaging domains, brain MRI suffers from high heterogeneity: multiple acquisition protocols, diverse sequence types (e.g., T1w, T2w, FLAIR, DWI), inconsistent annotations, fragmented repositories, and variable licensing terms. This fragmentation presents a unique challenge for developing general-purpose models.

A number of surveys and benchmarks have attempted to catalog medical imaging datasets, but they fall short in key ways when viewed from the perspective of brain MRI foundation models. For instance, MedSegBench \cite{kus_medsegbench_2024} aggregates 35 datasets across modalities, yet includes only one brain MRI dataset and provides no analysis of voxel-level heterogeneity or preprocessing standards. Similarly, Dishner et al.~\cite{dishner_survey_2024} catalogued 110 radiology datasets (49 brain MRI) but spanned too broad an anatomical scope to address brain-specific challenges such as sequence diversity and harmonization \cite{pomponio_harmonization_2020}. Other focused reviews, e.g., on glioma datasets \cite{yearley_current_2022, andaloussi_exploring_2024}, provide valuable clinical and molecular context but rarely analyze imaging metadata (e.g., voxel resolution, intensity distributions, or missing modalities) that directly influence pretraining strategies. Even highly influential initiatives like the BraTS Challenge \cite{menze_multimodal_2015, bakas_identifying_2018, baid_rsna-asnr-miccai_2021, bonato_advancing_2025} have advanced reproducibility and benchmarking but rely on heavily preprocessed data, which reduces heterogeneity and thus limits real-world generalization. In short, prior surveys tend to be either too broad (spanning many anatomical domains) or too narrow (focusing on a single disease), and they often omit the image- and preprocessing-level variability most relevant for foundation model development.

This review addresses these gaps. We provide a structured and multi-level assessment of public brain MRI datasets with a specific focus on their suitability for foundation model training. Unlike prior works, we move beyond cataloguing and explicitly quantify variability across dataset-level and image-level properties. We also evaluate the effects of preprocessing choices, which remain a largely underexplored source of covariate shift. Our analysis is designed to bridge the disconnect between dataset curation and model pretraining, highlighting practical considerations for building harmonized resources.

Our contributions are fourfold:
\begin{enumerate}[label=(\roman*), leftmargin=4em, align=left]
    \item Dataset-level review: We review \totaldatasets{} adult 3D structural brain MRI datasets covering over \totalsubjects{} subjects. This includes detailed analysis of modality composition, disease coverage, dataset scale, and licensing diversity, revealing major imbalances between healthy and clinical populations that influence pretraining data design.
    
    \item Image-level profiling: We perform a quantitative comparison of voxel spacing, image orientation, and intensity statistics across \selecteddatasets{} representative datasets. This analysis exposes strong variation in geometric resolution and contrast distribution, which can affect how foundation models learn anatomical and pathological features.
    
    \item Quantitative evaluation of preprocessing variability: We measure how bias field correction, intensity normalization, skull stripping, registration, and interpolation modify voxel-level statistics and geometry across datasets.
    
    \item Feature-space analysis of residual covariate shift: Using a 3D DenseNet121, we quantify cross-dataset divergence that remains after full preprocessing, linking voxel-level variability to learned representations.
\end{enumerate}

Together, these contributions provide the first structured review that unifies dataset-, image-, and preprocessing-level analyses, offering practical guidelines for building harmonized and generalizable brain MRI foundation models.

%% file: sections/methodology.tex
\section{Review Methodology}

\subsection{Data Collection and Selection Process}
We performed a structured search for publicly available brain MRI datasets between May and June 2025. Sources included Google, Google Dataset Search, PubMed, Scientific Data, and major neuroimaging repositories such as TCIA, OpenNeuro, NITRC, CONP Portal and Synapse. Search terms combined phrases such as ``public brain MRI dataset,'' ``open access brain MRI,'' ``3D structural brain MRI for AI,'' and ``MRI segmentation dataset,'' with variations replacing ``dataset'' by ``database.'' No date restrictions were applied. Each repository entry or publication was manually reviewed to determine eligibility, and the process was repeated iteratively until no new datasets were identified, achieving data saturation.

This review focused exclusively on datasets containing 3D structural MRI of the adult human brain. Datasets were included only if they satisfied all of the following criteria:

\begin{enumerate}[label=(\roman*), leftmargin=4em, align=left]
    \item volumetric 3D structural MRI scans were available (not 2D slices or statistical maps);
    \item subjects were adults;
    \item at least one structural modality (e.g., T1-weighted) was included, rather than only functional or diffusion modalities (e.g., fMRI, DTI, MRA);
    \item acquisitions were 3D static volumes (not 4D dynamic or time-resolved scans); and
    \item at least 20 unique 3D scans were provided.
\end{enumerate}

For multimodal datasets that additionally included fMRI, DTI, PET, or clinical assessments, only the structural MRI scans were considered in this review.  

\paragraph{Screening Outcome.}  
Our search yielded more than one hundred candidate entries across repositories and publications. After removing duplicates and excluding pediatric-only cohorts, 2D or statistical map datasets, collections with fewer than 20 scans, and datasets without accessible images, a total of 52 datasets were retained. Together, these cover 524{,}310 subjects and form the basis of our review.

\paragraph{Standardization of Modalities and Cohort Labels.}
To enable consistent comparison across heterogeneous datasets, we standardized both imaging modalities and cohort labels. 
The detailed mapping rules are summarized in Appendix Table~\ref{tab:modality_standardization} (modalities) and Table~\ref{tab:cohort_standardization} (cohorts).

These datasets span a broad range of neurological and psychiatric conditions alongside healthy controls, and vary in imaging protocols, scanner characteristics, and subject demographics. A complete overview is provided in Table~\ref{tab:total_datasets}.

\input{tables/total_datasets}

\paragraph{Subset for Image-level Analysis.}  
Due to licensing restrictions and regional access limitations, only a portion of the identified datasets could be downloaded for direct inspection. To avoid redundancy, we excluded benchmark collections that merely aggregate scans from other public sources, retaining only the original datasets. The subset used for image-level profiling includes: MSLesSeg \cite{guarnera_mslesseg_2025}, MS-60 \cite{ali_m_muslim_brain_2022}, MSSEG-2 \cite{miccai_2021_longitudinal_nodate}, BraTS25-MET \cite{maleki_analysis_2025}, BraTS25-SSA \cite{adewole_brain_2023, adewole_brats-africa_2025}, BraTS25-MEN \cite{labella_asnr-miccai_2023, labella_analysis_2024, labella_multi-institutional_2024}, ISLES22 \cite{hernandez_petzsche_isles_2022}, EPISURG \cite{perez-garcia_episurg_2020}, OASIS-1 \cite{marcus_open_2007}, OASIS-2 \cite{marcus_open_2010}, IXI \cite{noauthor_ixi_nodate}, UMF-PD \cite{badea_exploring_2017}, NFBS \cite{puccio_preprocessed_2016}, and BrainMetShare \cite{grovik_brainmetshare_2020}.  

\subsection{Metadata Extraction}
To enable consistent cross-dataset analysis, we programmatically loaded each image file and extracted key metadata. For every scan, we recorded spatial attributes (image dimensions, voxel spacing, orientation codes, affine matrix) and non-image attributes (modality, subject ID, session ID when available). Images outside the inclusion scope, such as DTI sequences in IXI, were excluded at this stage.  

All extracted metadata were stored in standardized per-dataset CSV files following a uniform schema. This structured resource forms the foundation for subsequent dataset- and image-level analyses presented in this review and is designed to facilitate reproducibility and reuse by the wider community.

%% file: tables/total_datasets.tex
\renewcommand{\arraystretch}{1.2}

\begin{longtable}{p{4cm} p{3.75cm} p{3.75cm} p{2cm}}
\caption{Summary of included brain MRI datasets.} \\
\label{tab:total_datasets} \\
\hline
\textbf{Dataset} & \textbf{Modality} & \textbf{Cohort} & \textbf{\#Subjects} \\
\hline
\endfirsthead

\multicolumn{4}{c}{{\tablename\ \thetable{} -- continued from previous page}} \\
\hline
\textbf{Dataset} & \textbf{Modality} & \textbf{Cohort} & \textbf{\#Subjects} \\
\hline
\endhead

\hline \multicolumn{4}{r}{{Continued on next page}} \\
\endfoot

\hline
\endlastfoot

{[18F]MK6240} \cite{dascal_open_2025} & T1, Others & Healthy & 33 \\
ABIDE-I \cite{di_martino_autism_2014} & T1 & Autism, Healthy & 1,112 \\
ABIDE-II \cite{di_martino_enhancing_2017} & T1 & Autism, Healthy & 1,114 \\
ADNI \cite{jack_alzheimers_2008, jack_overview_2024} & T1, T2, FLAIR & Neurodegenerative & 4,068 \\
AOMIC-ID1000 \cite{snoek_aomic-id1000_2021, snoek_amsterdam_2021} & T1, DWI/DTI, fMRI/rs-fMRI & Healthy & 928 \\
AOMIC-PIOP1 \cite{snoek_aomic-piop1_2020, snoek_amsterdam_2021} & T1, DWI/DTI, fMRI/rs-fMRI & Healthy & 216 \\
AOMIC-PIOP2 \cite{snoek_aomic-piop2_2020, snoek_amsterdam_2021} & T1, DWI/DTI, fMRI/rs-fMRI & Healthy & 226 \\
ARC \cite{gibson_aphasia_2023, gibson_aphasia_2024} & T1, T2, FLAIR, DWI/DTI, fMRI/rs-fMRI & Stroke & 230 \\
ATLAS R2.0 \cite{liew_large_2022} & T1 & Stroke & 955 \\
BBSRC \cite{lloyd_emotion_2021} & T1, DWI/DTI, fMRI & Healthy & 34 \\
BrainMetShare \cite{grovik_brainmetshare_2020} & T1, T1C, FLAIR & Brain Tumor & 156 \\
BraTS-GLI (2025) \cite{verdier_2024_2024, baid_rsna-asnr-miccai_2021} & T1, T1C, T2, FLAIR & Brain Tumor & 1,809 \\
BraTS-MEN (2025) \cite{labella_asnr-miccai_2023, labella_analysis_2024, labella_multi-institutional_2024} & T1, T1C, T2, FLAIR & Brain Tumor & 750 \\
BraTS-MET (2025) \cite{maleki_analysis_2025} & T1, T1C, T2, FLAIR & Brain Tumor & 1,778 \\
BraTS-SSA (2025) \cite{adewole_brain_2023, adewole_brats-africa_2025} & T1, T1C, T2, FLAIR & Brain Tumor & 95 \\
CC-359 \cite{souza_open_2018} & T1 & Healthy & 359 \\
DLBS \cite{park_dallas_2024} & T1, T2, FLAIR, DWI/DTI, fMRI/rs-fMRI & Healthy & 464 \\
EDEN2020 \cite{castellano_eden2020_2020} & T1C, FLAIR, DWI/DTI, Others & Brain Tumor, Healthy & 45 \\
EPISURG \cite{perez-garcia_episurg_2020} & T1 & Epilepsy & 430 \\
GSP \cite{buckner_brain_2014} & T1, DWI/DTI, fMRI/rs-fMRI & Healthy & 1,570 \\
HBN-SSI \cite{oconnor_healthy_2017} & T1, Others & Healthy & 13 \\
HCP \cite{van_essen_wu-minn_2013} & T1, fMRI/rs-fMRI & Healthy & 1,200 \\
ICTS \cite{lu_intracranial_2021} & T1C & Brain Tumor & 1,591 \\
IDB-MRXFDG \cite{merida_cermep-idb-mrxfdg_2021} & T1, FLAIR, Others & Healthy & 37 \\
IDEAS \cite{taylor_imaging_2024} & T1, FLAIR & Epilepsy & 442 \\
ISLES22 \cite{hernandez_petzsche_isles_2022} & T1, T2, DWI/DTI, FLAIR & Stroke & 400 \\
IXI \cite{noauthor_ixi_nodate} & T1, T2, DWI/DTI & Healthy & 581 \\
MBSR \cite{seminowicz_mbsr_2024} & T1, DWI/DTI, fMRI & Healthy & 147 \\
Brain Tumor-SEG-CLASS \cite{vassantachart_segmentation_2023} & T1, T1C, FLAIR & Brain Tumor & 96 \\
MGH Wild \cite{iglesias_synthsr_2023} & T1, T2, FLAIR & Healthy & 1,110 \\
MICA-MICs \cite{royer_open_2022} & T1, DWI/DTI, fMRI/rs-fMRI & Healthy & 50 \\
MOTUM \cite{zhenyu_gong_multi-center_2023} & T1, T1C, T2, FLAIR & Brain Tumor & 66 \\
MS-60 \cite{ali_m_muslim_brain_2022} & T1, T2, FLAIR & Multiple Sclerosis & 60 \\
MSLesSeg \cite{guarnera_mslesseg_2025} & T1, T2, FLAIR & Multiple Sclerosis & 75 \\
MSSEG-2 \cite{miccai_2021_longitudinal_nodate} & FLAIR & Multiple Sclerosis & 100 \\
MSValid \cite{pappalardo_msvalid_2024} & T1, T2, FLAIR & Multiple Sclerosis & 84 \\
NFBS \cite{puccio_preprocessed_2016} & T1 & Psychiatric Disorders, Healthy & 125 \\
NIMH-Ketamine \cite{evans_nimh_2025} & T1, T2, DWI/DTI, fMRI & Psychiatric Disorders, Healthy & 58 \\
NIMH-RV \cite{nugent_nimh_2025, nugent_nimh_2022} & T1, T2, DTI, FLAIR, Others & Healthy & 1,859 \\
Novosibirsk-Brain Tumor \cite{filimonova_utilizing_2024} & T2, FLAIR, DWI/DTI, Others & Brain Tumor & 42 \\
OASIS-1 \cite{marcus_open_2007} & T1 & Neurodegenerative & 416 \\
OASIS-2 \cite{marcus_open_2010} & T1 & Neurodegenerative & 150 \\
PPMI \cite{marek_parkinsons_2018} & CT, fMRI, MRI, DTI, PET, SPECT & Neurodegenerative, Healthy & 8,765 \\
QIN-BRAIN-DSC-MRI \cite{schmainda_glioma_2016} & T1, DSC & Brain Tumor & 49 \\
ReMIND \cite{juvekar_brain_2023} & T1, T1C, T2, FLAIR, DWI/DTI, iUS & Brain Tumor & 114 \\
SOOP \cite{rorden_stroke_2024} & T1, T2, FLAIR, TRACE, ADC & Stroke & 1,669 \\
UCLA \cite{bilder_ucla_2020, poldrack_phenome-wide_2016} & T1, DWI/DTI, fMRI/rs-fMRI & Psychiatric Disorders, Healthy & 272 \\
UCSF-ALPTDG \cite{fields_university_2024} & FLAIR, T1, T1C, T2 & Brain Tumor & 298 \\
UCSF-BMSR \cite{rudie_university_2024} & T1, T1C, FLAIR & Brain Tumor & 412 \\
UCSF-PDGM \cite{calabrese_university_2023} & T1, T1C, T2, FLAIR, DWI/DTI, Others & Brain Tumor & 495 \\
UKBioBank \cite{sudlow_uk_2015} & T1, FLAIR, DWI/DTI, fMRI/rs-fMRI & Multiple Diseases & 500,000 \\
UMF-PD \cite{badea_exploring_2017} & T1, fMRI/rs-fMRI & Neurodegenerative, Healthy & 83 \\
UPENN-GBM \cite{bakas_multi-parametric_2021, bakas_university_2022} & T1, T1C, T2, FLAIR, Others & Brain Tumor & 630 \\
WMH \cite{kuijf_data_2022} & T1, FLAIR & White Matter Hyperintensities & 170 \\
\end{longtable}

%% file: sections/dataset_level.tex
\section{Dataset-Level Analysis}

\subsection{Disease Coverage}
\begin{figure}[h]
    \centering
    \includegraphics[width=0.8\textwidth]{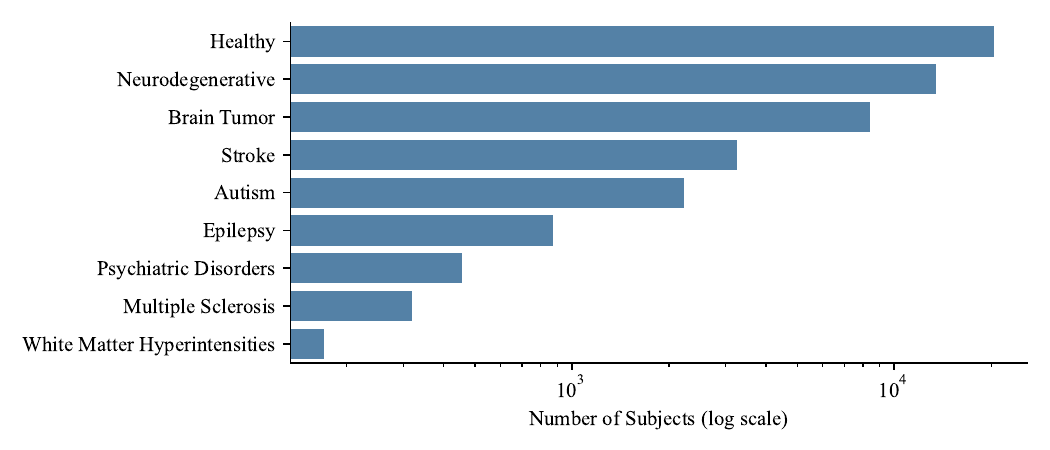}
    \caption{Distribution of subjects by disease category after removing the undefined ``Multiple Diseases'' group. The x-axis uses a logarithmic scale to enable visualization across several orders of magnitude, from hundreds to tens of thousands of subjects.}
    \label{fig:disease_distribution}
\end{figure}

The disease distribution analysis shown in Figure~\ref{fig:disease_distribution} reveals a pronounced imbalance across public brain MRI datasets. After separating combined cohort labels and removing the undefined “Multiple Diseases” category, \textit{Healthy} subjects form the largest group, followed by \textit{Neurodegenerative disorders} (approximately 8,800 subjects) and \textit{Brain Tumors} (around 8,400 subjects). Medium-scale categories include \textit{Stroke} (2,300 subjects), \textit{Autism} (2,200 subjects), and \textit{Epilepsy} (870 subjects). Smaller datasets correspond to \textit{Psychiatric Disorders} (455 subjects), \textit{Multiple Sclerosis} (319 subjects), and \textit{White Matter Hyperintensities} (170 subjects). 

This distribution highlights the structural bias of the open neuroimaging landscape. The abundance of healthy and neurodegenerative cohorts reflects the historical focus on population-based and aging studies, while chronic, diffuse, or subtle pathologies remain underrepresented. Despite the diversity of available datasets, the dominance of a few diagnostic categories implies that current public MRI data cannot fully capture the clinical heterogeneity of the brain. This skewed representation constrains comparative analysis across disease types and may perpetuate overrepresentation of high-resource conditions in future benchmarks.

For foundation models, the imbalance in disease coverage directly influences representational learning. Pretraining dominated by T1-weighted healthy and Alzheimer’s data encourages the model to learn structural regularities and global contrast variations, while subtle lesion characteristics typical of demyelinating or vascular diseases remain statistically rare. Such bias limits transferability to small-lesion or microstructural disorders. To mitigate this, pretraining datasets should deliberately balance disease composition, incorporate underrepresented conditions (e.g., MS, WMH, psychiatric disorders), and include healthy scans primarily as anatomical anchors. Transparent reporting of disease proportions is essential for understanding bias propagation during large-scale pretraining.

\subsection{Dataset Scale}

\begin{figure}[h]
    \centering
    \includegraphics[width=0.75\textwidth]{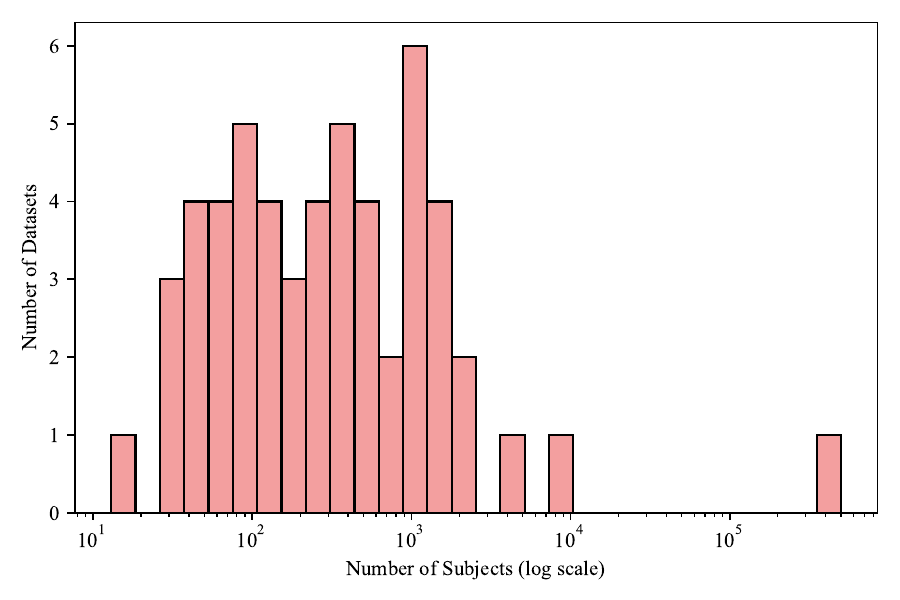}
    \caption{Distribution of dataset sizes on a logarithmic scale. The figure highlights the dominance of one extremely large population dataset and numerous smaller, clinically focused cohorts. Logarithmic scaling compresses large numerical differences to emphasize structural imbalance across dataset scales.}
    \label{fig:dataset_scale}
\end{figure}

The analysis of dataset sizes (Figure~\ref{fig:dataset_scale}) exposes an extreme imbalance in the public brain MRI landscape. A single dataset—UKBioBank—accounts for more than 500,000 subjects, while nearly all other datasets range from a few dozen to a few thousand participants. Yet when examined alongside disease coverage, the relationship between scale and content becomes more revealing: the largest datasets are almost exclusively composed of healthy or aging populations, whereas smaller datasets concentrate on specific pathologies such as brain tumors, stroke, and multiple sclerosis. In other words, data abundance is inversely correlated with clinical complexity.
 
For foundation models, the insight from this scale–disease relationship is profound. Pretraining must not simply accumulate images—it must balance information density against population scale. Large healthy datasets can anchor the model’s low-level feature representation, but meaningful generalization arises only when smaller, heterogeneous clinical datasets are interleaved to inject structural variability and abnormal morphology. The optimal training corpus is therefore not the largest one, but the one that combines datasets across scales and disease domains in a way that maximizes representational complementarity. 

When merging datasets, several considerations follow:
\begin{itemize}
    \item Sampling balance: naive aggregation will cause population-scale datasets to dominate optimization; adaptive weighting or stratified sampling is necessary to preserve rare clinical features.
    \item Harmonization: resolution, voxel spacing, and intensity normalization must be aligned to prevent the model from interpreting acquisition differences as anatomical variations.
    \item Domain alignment: cross-dataset normalization in feature space (e.g., domain-adversarial training or latent alignment) can reduce the domain gap between healthy and disease cohorts.
\end{itemize}

The scale analysis reveals that the most informative foundation model will not come from the largest dataset, but from the strategic fusion of small, diverse datasets with large, stable ones. Quantity establishes the foundation; diversity defines intelligence. A model pretrained under this philosophy learns both the invariant anatomy of the healthy brain and the variable morphology of disease, achieving robustness not through volume, but through representational balance.

\subsection{Modality Composition}
\begin{figure}[h]
    \centering
    \includegraphics[width=0.75\textwidth]{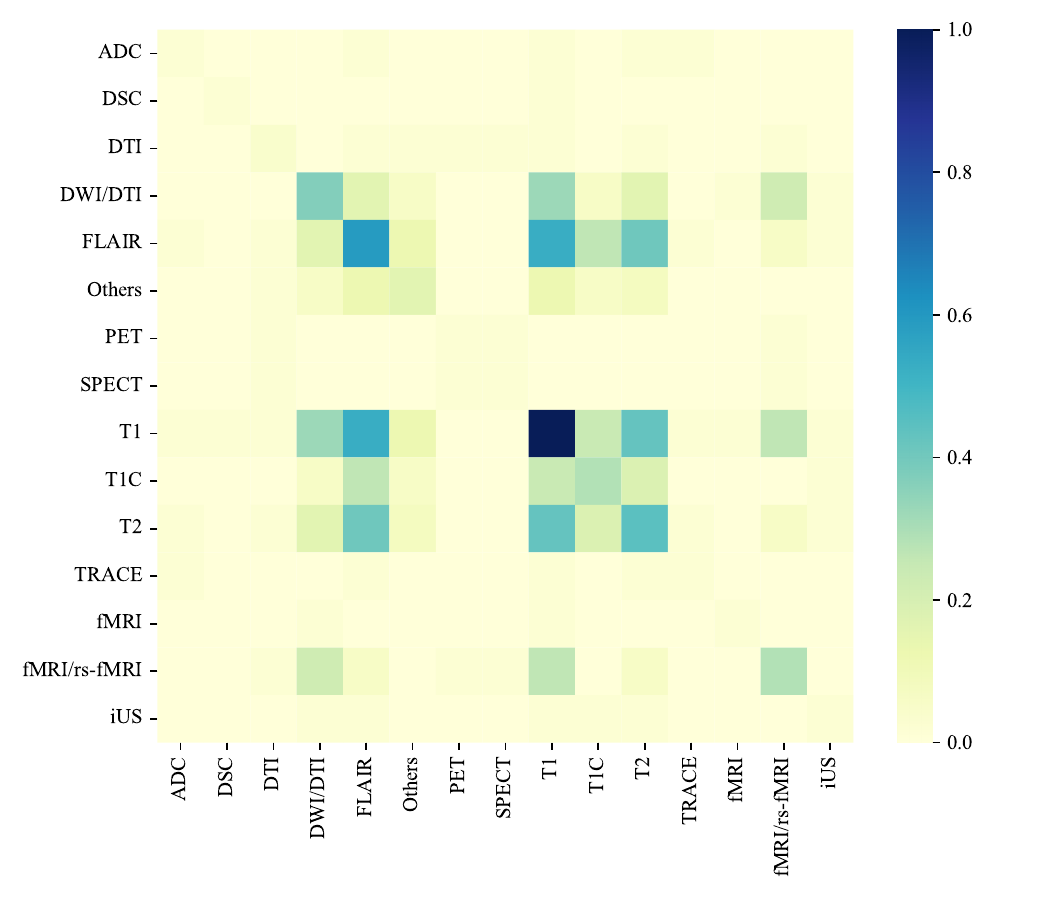}
    \caption{Heatmap of modality co-occurrence across structural MRI datasets. High-intensity cells indicate frequent pairing between modalities, particularly T1–FLAIR, T1–T2, and T1–T1c. These patterns reveal partial but consistent overlap that supports unified representation learning across multi-dataset collections.}
    \label{fig:modality_overlap}
\end{figure}

The modality co-occurrence analysis (Figure~\ref{fig:modality_overlap}) reveals distinct pairing patterns among structural MRI sequences across public datasets. The most frequent combination is between T1-weighted and FLAIR scans, followed by T1–T2 and T1–T1c pairs. These sequences commonly co-occur within multi-contrast structural datasets such as BraTS, ADNI, and MSSEG, where complementary contrasts are used to capture both anatomical boundaries and pathological hyperintensities. Moderate co-occurrence is also observed among FLAIR, T2, and T1c, indicating a tendency for lesion-focused studies to integrate multiple structural contrasts that highlight different tissue characteristics. In contrast, single-modality datasets remain prevalent, particularly among population studies (e.g., IXI, OASIS), which provide only T1-weighted scans.

This co-occurrence pattern demonstrates that public brain MRI datasets—though diverse—are structurally interlinked through a limited but consistent set of core modalities. The strong correlation between T1 and FLAIR availability reflects a shared acquisition strategy for anatomical delineation and lesion sensitivity, while the partial inclusion of T2 and T1c indicates dataset-specific clinical emphasis (e.g., edema or contrast enhancement). The heatmap also reveals that cross-dataset modality overlap is incomplete: no single dataset provides full structural coverage, and different combinations dominate different disease domains. This partial alignment introduces redundancy in some modalities but gaps in others when datasets are combined.

For foundation models trained on aggregated public datasets, these co-occurrence dynamics carry important consequences. The uneven intersection of modalities across datasets means that multi-contrast information is not uniformly available for all subjects. This heterogeneity can lead to modality imbalance during pretraining and complicate cross-dataset harmonization. To address this, foundation models must incorporate modality-aware mechanisms—such as learned modality embeddings or masked reconstruction objectives—that can leverage overlapping contrasts while remaining robust to missing ones. The observed co-occurrence structure also suggests that structural modalities share sufficient anatomical redundancy to enable joint representation learning: by training across datasets with partially overlapping contrasts (e.g., T1+FLAIR from one source, T1+T2 from another), the model can implicitly learn a unified structural feature space that generalizes across acquisition protocols. Consequently, modality co-occurrence is not merely a dataset property but a key enabler of scalable, harmonized pretraining across heterogeneous MRI corpora.

%% file: sections/image_level.tex
\section{Image-Level Analysis}
At the image level, heterogeneity in voxel geometry, orientation, and intensity introduces latent biases that can substantially affect representation learning. These properties define the physical scale, spatial consistency, and dynamic range of brain MRI data—factors that determine whether a foundation model learns anatomical invariants or dataset-specific artifacts. Our image-level analysis quantifies these factors across 14 public datasets and provides interpretative insights for model design and harmonization.

\subsection{Voxel Spacing}

Voxel spacing defines the physical size of each voxel along the $x$, $y$, and $z$ axes in millimeters, determining how finely anatomical structures are represented in the image and directly influencing the learning behavior of foundation models. When voxel spacing varies across datasets, the same convolution or attention kernel covers different physical regions, leading to inconsistent representation of anatomical details, blurred or missing small lesions in thicker slices, and domain shifts when combining data. This makes voxel spacing not just a technical aspect of MRI acquisition but a key factor that shapes model generalization. It affects architectures differently: CNNs may learn biased features when scale changes, transformers can misalign patches or positional encodings, and SAM-style models often lose boundary accuracy when slices are uneven—making anisotropy a hidden source of error that limits transferability.

\begin{figure}[h]
\centering
\includegraphics[scale=0.8]{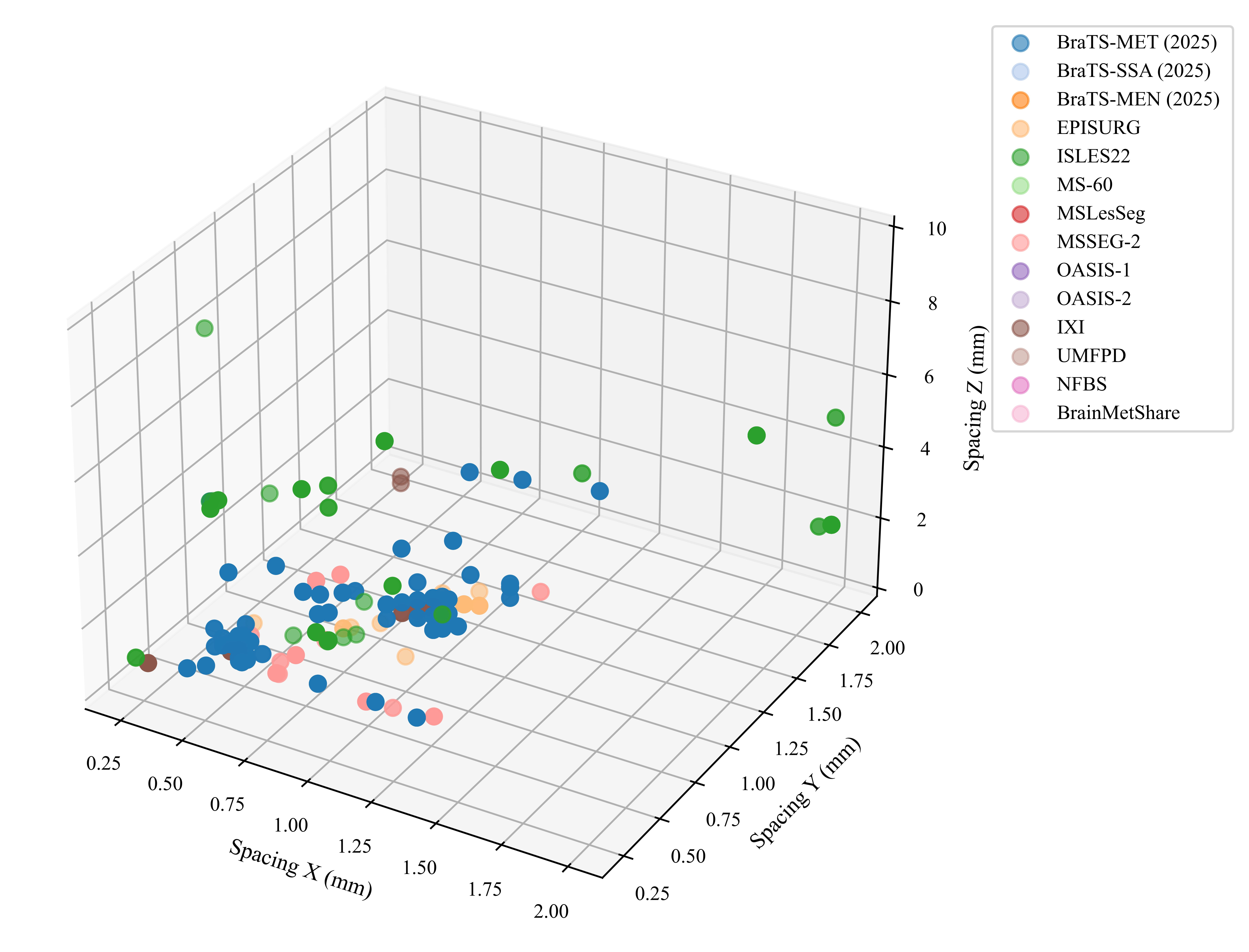}
\caption{Voxel spacing distribution (in mm) along the $x$, $y$, and $z$ axes for 14 curated datasets. Each point represents one scan, and each color corresponds to a dataset. Compact clusters indicate consistent acquisition protocols, while spread-out points show variation in resolution and anisotropy.}
\label{fig:voxel_3d}
\end{figure}

Figure~\ref{fig:voxel_3d} shows the 3D distribution of voxel spacings across 14 representative datasets. Most datasets cluster near isotropic spacing around $(1.0, 1.0, 1.0)$ mm, indicating uniform resolution across all axes. The three BraTS collections (BraTS-MET, BraTS-SSA, BraTS-MEN), OASIS-1/2, NFBS, and IXI fall into this group, providing consistent high-quality data for model pretraining.  
In contrast, multiple sclerosis datasets (MS-60, MSLesSeg, MSSEG-2) and BrainMetShare exhibit moderate anisotropy, with fine in-plane resolution ($x,y \approx 0.8{-}1.0$ mm) but thicker slices along the $z$-axis ($1.5{-}3.0$ mm). This reduces sensitivity to small or thin lesions that appear across only one or two slices.  
Stroke and surgical datasets, such as ISLES22 and EPISURG, show the widest variability, including cases with very thick slices ($z>4$ mm) and variable in-plane spacing up to $2.0$ mm. Such heterogeneity reflects differences in acquisition protocols across centers and scanners. Finally, mixed clinical datasets like UMFPD and BrainMetShare include both near-isotropic and anisotropic scans, representing real-world diversity in clinical imaging practices.
These observations lead to three key insights that have direct implications for the development of foundation models: (i) Many research datasets share near-isotropic resolution and are well-suited for standardized pretraining; (ii) Clinical and disease-specific datasets tend to be anisotropic, introducing geometric inconsistencies that require explicit modeling; and (iii) spacing variability alone can cause measurable distribution shifts between datasets, even after resampling.

To further characterize these differences, we grouped each image into three categories based on the degree of anisotropy. We computed, for each image, the ratio between the largest and smallest spacing values among the three axes. If all spacings were equal (ratio = 1.0), the image was labeled as \textit{isotropic}. If the ratio was greater than 1.0 but less than 2.0, it was labeled \textit{mildly anisotropic}. Ratios of 2.0 or higher were labeled \textit{highly anisotropic}.

\input{tables/anisotropy}

As shown in Table~\ref{tab:anisotropy}, most images fall into the isotropic or mildly anisotropic categories—approximately 7,968 and 7,152 images, respectively. However, over 1,700 images are highly anisotropic, indicating substantial geometric distortion, especially in slice thickness. If left uncorrected, these differences can lead to biased model learning and performance degradation across datasets.

\subsection{Orientation}

The orientation of MRI volumes defines how the anatomical axes of the brain are mapped to the voxel coordinate system. Each MRI scan stores its orientation using a three-letter code (e.g., RAS, LAS, LPS), which specifies the direction of the \textit{x}, \textit{y}, and \textit{z} axes relative to the patient’s anatomy. While orientation may appear as a technical metadata field, it has a direct and critical influence on the learning behavior of foundation models. When images are stored in inconsistent orientations across datasets, identical brain structures appear in different spatial locations or mirrored configurations. This leads to misalignment in anatomical correspondences, causing the model to learn orientation-specific patterns instead of generalizable anatomical features. Therefore, harmonizing orientation is essential for foundation models to learn consistent spatial representations that can generalize across diverse datasets.
 
Table~\ref{tab:axcodes_distribution} summarizes the orientation distribution across datasets. The most common orientation is RAS (6,592 images), which is the standard convention in neuroimaging software such as FSL and FreeSurfer. However, a considerable number of datasets adopt alternative conventions, including LPS (5,012 images) and LAS (3,473 images). These three orientations together account for over 90\% of all images analyzed. Notably, several datasets contain multiple orientations internally—for instance, BraTS-MET and EPISURG each include images in both RAS and LPS forms. Less frequent orientations such as RSA, PSR, or ASL are observed in smaller datasets (e.g., OASIS, NFBS, UMFPD). The presence of such variability reflects the absence of a unified orientation policy among dataset providers, even within well-curated public repositories.

\input{tables/axcodes_distribution}

The observed orientation heterogeneity introduces a subtle but significant source of distributional shift that can impair model transferability. Models trained on mixed-orientation data without explicit normalization may implicitly encode orientation-specific spatial priors. For example, left–right inversions between RAS and LAS orientations can confuse the model’s learned feature alignment, leading to inconsistent activation patterns for homologous brain regions. Similarly, inconsistent superior–inferior axis definitions can distort 3D spatial context, reducing the model’s ability to capture global anatomical symmetry.  

For foundation model pretraining, these inconsistencies compound across large-scale datasets. Since pretraining relies on learning generic spatial and structural representations, uncorrected orientation differences can fragment the learned latent space, -0
causing the model to associate the same anatomy with distinct feature embeddings depending on orientation. This weakens the universality of learned representations and increases the burden on fine-tuning.  

Hence, orientation harmonization is not merely a preprocessing detail but a foundational requirement for effective cross-dataset learning. Converting all volumes to a common convention (typically RAS) before model training ensures that spatial relationships are consistent across datasets. For large-scale pretraining pipelines, we recommend enforcing explicit orientation standardization as part of dataset ingestion. Such harmonization minimizes unnecessary domain shifts, allowing the foundation model to focus on learning biologically meaningful anatomy rather than orientation artifacts.

\subsection{Image Intensity Distribution}

Image intensity represents the voxel-wise signal values within MRI scans and encapsulates the physical properties of tissues as captured by different imaging sequences. Intensity distributions are shaped by scanner hardware, acquisition protocols, and post-processing pipelines such as bias-field correction or 
intensity normalization. For foundation models, which depend on large-scale data aggregation from diverse sources, inconsistent intensity scaling or contrast 
profiles can substantially affect representation learning. A model trained on non-harmonized intensity profiles may implicitly overfit to dataset-specific 
brightness ranges, thereby reducing its ability to generalize across unseen domains.

\begin{figure}[h]
    \centering
    \includegraphics[width=\textwidth]{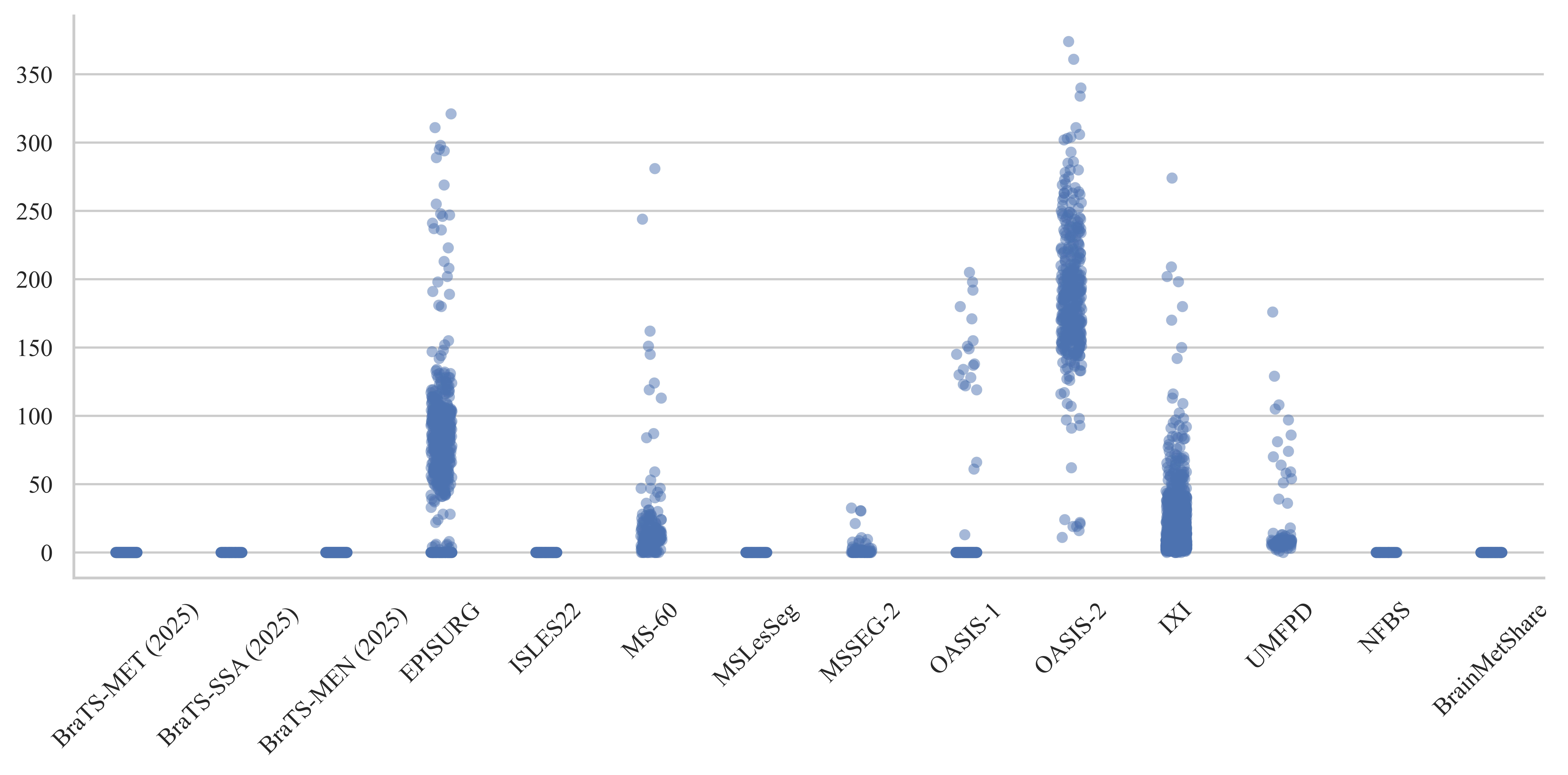}
    \caption{Median voxel intensity per image across datasets. Each dot 
    represents one 3D MRI volume.}
    \label{fig:median_intensity}
\end{figure}

Figure~\ref{fig:median_intensity} illustrates the distribution of median voxel intensities across representative datasets. Datasets such as EPISURG, OASIS-1, OASIS-2, and IXI exhibit wide intensity variability, whereas others (e.g., the BraTS series, ISLES22, MSLesSeg, and BrainMetShare) show lower and more stable median values. This disparity likely arises from differences in scanner calibration, rescaling conventions (e.g., 0–255 versus z-scored), and preprocessing intensity normalization methods. The OASIS datasets, for example, show extensive dispersion with median intensities exceeding 300, reflecting a broad dynamic range and the absence of uniform scaling. In contrast, the BraTS and MS-related datasets exhibit tight clusters around zero, suggesting that bias correction and standardized normalization were consistently applied.

These differences have several implications for foundation model development. First, heterogeneous intensity distributions introduce latent biases that may lead a model to associate tissue contrast with dataset identity rather than underlying anatomy. This undermines the objective of learning scanner- and modality-invariant representations. Second, extreme intensity outliers—particularly in datasets with mixed acquisition conditions—can destabilize loss optimization during pretraining by distorting the input statistics used by normalization layers. Conversely, datasets with highly standardized intensity ranges, while beneficial for stable convergence, may limit the model’s exposure to real-world variability and thus reduce robustness during fine-tuning on unnormalized clinical data.

From a model design perspective, these findings highlight the importance of preprocessing-aware normalization strategies. Dynamic intensity scaling or adaptive histogram alignment could be implemented within the data loading pipeline to ensure consistent contrast across datasets. Alternatively, self-supervised objectives that promote intensity-invariant representations (e.g., histogram-matching augmentations or contrast consistency losses) may help the model decouple anatomical features from brightness variations. Ultimately, balancing intensity harmonization for stable training with sufficient distributional diversity for adaptability remains a key challenge for developing robust and generalizable MRI foundation models.

To quantitatively assess whether these intensity differences are statistically significant, we applied the Kruskal–Wallis H test to the per-image median values grouped by dataset. The result was highly significant ($H = 15093.849$, $p < 0.0001$), confirming that the observed inter-dataset variations are not due to random fluctuation. This non-parametric test evaluates whether at least one 
group differs in median from the others, without assuming a specific underlying distribution. The extremely low $p$-value supports the visual findings in Figure~\ref{fig:median_intensity}, indicating that intensity scaling differences across datasets are real, systematic, and substantial.

%% file: tables/anisotropy.tex
\begin{table}[h]
\centering
\caption{Counts of anisotropy categories across all analyzed images.}
\label{tab:anisotropy}
\begin{tabular}{l r}
\toprule
\textbf{Anisotropy Category} & \textbf{Count} \\
\midrule
Isotropic & 7,968 \\
Mildly Anisotropic & 7,152 \\
Highly Anisotropic & 1,724 \\
\bottomrule
\end{tabular}
\end{table}

%% file: tables/axcodes_distribution.tex
\begin{table}[h]
\centering
\small
\begin{tabular}{l r p{9cm}}
\toprule
\textbf{Orientation} & \textbf{Count} & \textbf{Datasets} \\
\midrule
RAS & 6,592 & BraTS-MET (2025), BraTS-SSA (2025), BrainMetShare, EPISURG, ISLES22 \\
LPS & 5,012 & BraTS-MEN (2025), BraTS-MET (2025), BraTS-SSA (2025) \\
LAS & 3,473 & BraTS-MET (2025), EPISURG, ISLES22, IXI, MS-60, MSLesSeg, MSSEG-2, OASIS-1, UMFPD \\
RSA & 664 & EPISURG \\
PSR & 582 & EPISURG, IXI \\
ASL & 373 & OASIS-2 \\
PIR & 129 & EPISURG, NFBS \\
LIP & 9 & EPISURG \\
LSP & 5 & EPISURG \\
ASR & 4 & EPISURG \\
LSA & 1 & EPISURG \\
\bottomrule
\end{tabular}
\caption{Axial orientation distribution and dataset sources. Each axcode string represents the anatomical direction of the image axes: the first letter indicates the direction of the X-axis (e.g., R = Right, L = Left), the second letter corresponds to the Y-axis (e.g., A = Anterior, P = Posterior), and the third letter represents the Z-axis (e.g., S = Superior, I = Inferior). For example, RAS means the X-axis increases from left to right, the Y-axis from posterior to anterior, and the Z-axis from inferior to superior—commonly used in neuroimaging.}
\label{tab:axcodes_distribution}
\end{table}

%% file: sections/preprocessing.tex
\section{Evaluation of Preprocessing Effects on Image Harmonization}

To systematically evaluate the impact of preprocessing on data harmonization, we randomly sampled
images from the curated datasets and applied a standardized pipeline comprising bias-field correction,
intensity normalization, skull stripping, and spatial registration. The resulting images were analyzed
through voxel-wise statistical comparisons and qualitative visual inspection to assess improvements
in inter-dataset consistency and anatomical fidelity.

\subsection{Intensity Normalization}

Intensity normalization is the process of adjusting MRI voxel values to a common scale so that images from different scanners or subjects become comparable. The most common techniques include z-score normalization, histogram matching, and WhiteStripe normalization. Z-score normalization rescales each image to have zero mean and unit variance, reducing intensity range differences; it is best used as a simple, general method when datasets are diverse or lack a consistent reference. Histogram matching aligns the intensity distribution of each image to that of a reference scan or template, making it ideal for multi-site datasets with large scanner or protocol variability. WhiteStripe normalization uses the intensity range of normal-appearing white matter to anchor scaling, which is most effective for brain studies where maintaining tissue contrast is important.

\begin{figure}[h]
\centering
\includegraphics[width=\linewidth]{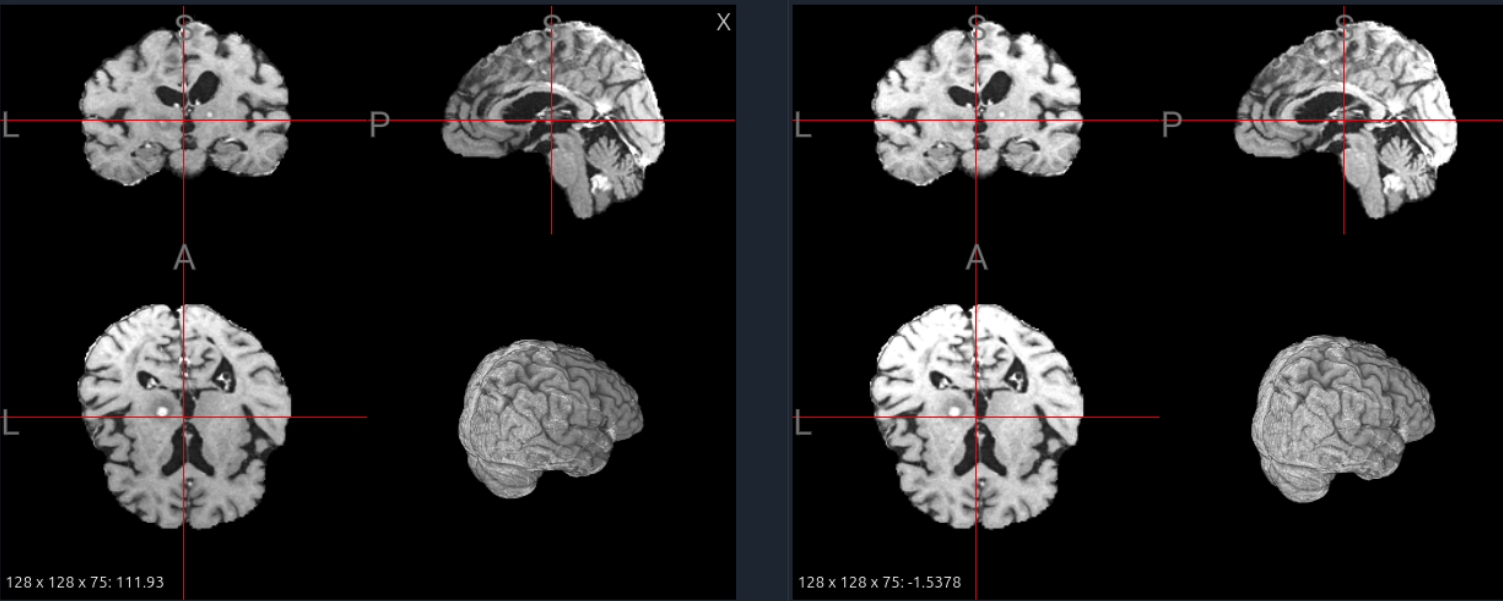}
\caption{T1C image from the BrainMetShare dataset before (left) and after z-score normalization (right).}
\label{fig:zscore_brainmetshare}
\end{figure}

As summarized in Table~\ref{tab:intensity_stats}, the original voxel intensities span a wide range, reflecting strong contrast between bright enhancement regions and darker tissues. After applying z-score normalization, the intensity distribution becomes centered around zero with reduced variance, resulting in a more uniform and balanced appearance across tissues. However, this transformation also alters the visual contrast, as shown in Figure~\ref{fig:zscore_brainmetshare}: some brain regions appear brighter, while fine structural details become less pronounced. This effect occurs because z-score normalization rescales voxel values relative to the global mean and standard deviation, thereby compressing the overall dynamic range and reducing intensity extremes.

When building foundation models, intensity normalization should be applied consistently across all datasets to prevent artificial domain shifts. The chosen method must preserve relative tissue contrast while harmonizing global intensity ranges. It is also beneficial to expose the model to multiple normalization styles during pretraining, helping it learn invariance to contrast variations. Finally, combining preprocessing-based normalization with learnable normalization layers (e.g., instance or adaptive layer normalization) allows the model to adapt dynamically to unseen data while maintaining stable, harmonized feature representations.

\input{tables/pp_intensity}

\subsection{Bias Field Correction}

Bias field correction adjusts MRI images to remove gradual brightness variations caused by uneven magnetic fields or coil sensitivity. These variations make some regions look brighter or darker even when the tissue is the same, so correction helps make the intensity more uniform across the brain. The popular methods include N4ITK (N4 bias field correction), N3 (nonparametric nonuniform intensity normalization), and SPM’s unified segmentation approach. In this review, we applied N4ITK bias correction with modality-specific tuning, such as adjustments included enhanced smoothing for FLAIR, brain masking for T1C images, and balanced settings for T1 and T2, using the SimpleITK implementation.

\begin{figure}[h]
    \centering
    \includegraphics[width=\linewidth]{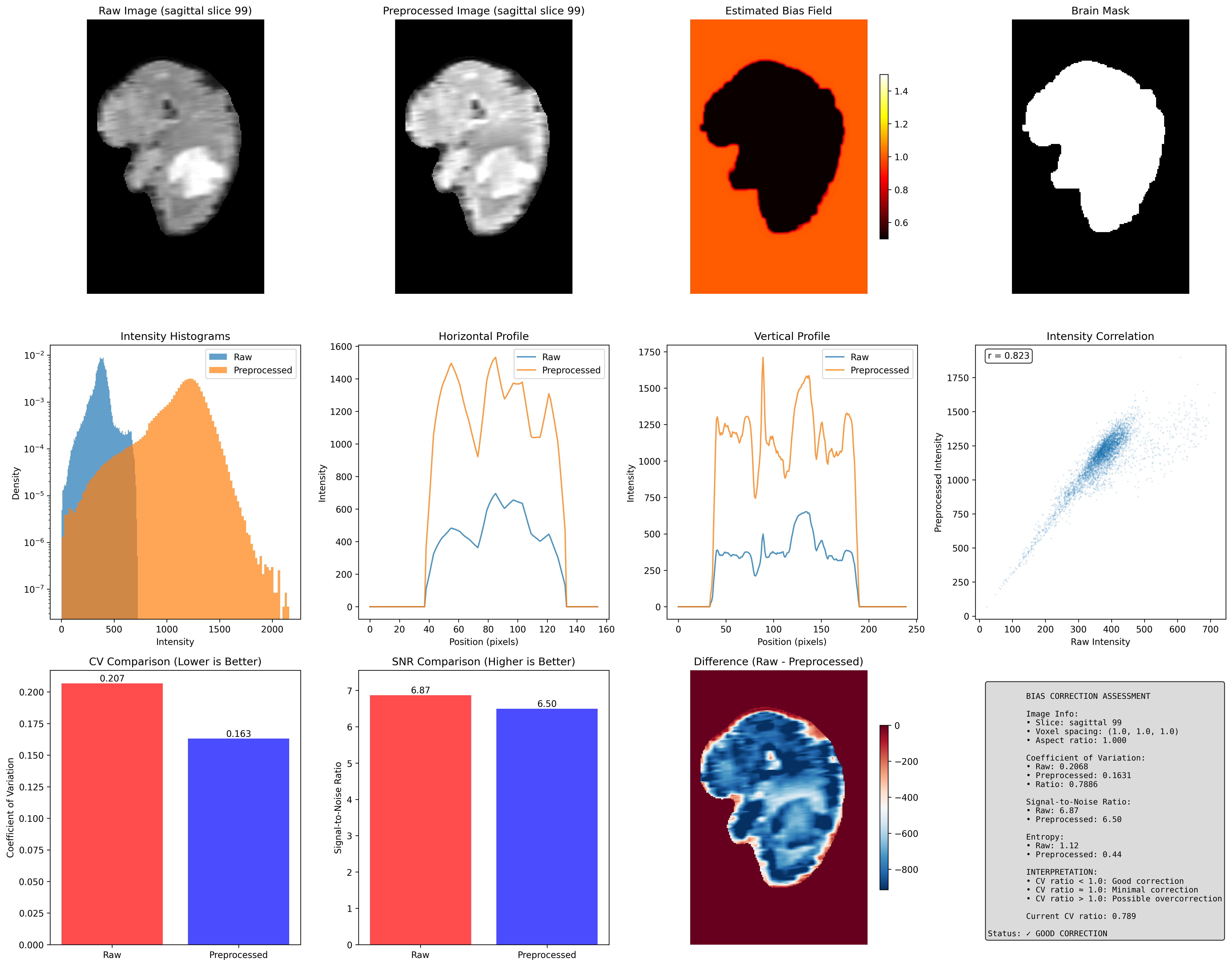}
    \caption{Bias field correction effect for BraTS-SSA. Includes raw image, corrected result, estimated bias field map, and applied brain mask.}
    \label{fig:bfc_brats_ssa}
\end{figure}

Representative examples are shown in Figure~\ref{fig:bfc_brats_ssa}. The raw image (top-left) displays uneven brightness — the left side of the brain appears darker due to scanner-related field inhomogeneity. After correction (top-second), the preprocessed image shows more uniform brightness across tissue regions, while the estimated bias field map (top-third) captures the smooth multiplicative field responsible for this nonuniformity. The intensity histograms reveal that voxel intensities have shifted and become more compact, indicating reduced variation between bright and dark areas. The horizontal and vertical profiles show that peaks corresponding to white matter and gray matter are now closer in amplitude, confirming improved intensity consistency. The intensity correlation plot (r = 0.823) shows that the correction maintains overall intensity relationships but rescales them toward a more uniform distribution. Quantitatively, as shown in Table~\ref{tab:bfc_stats}, the coefficient of variation decreases (0.207 → 0.163), meaning intensity variability within tissue is reduced, while the signal-to-noise ratio (SNR) remains similar (6.87 → 6.50), suggesting correction did not distort contrast or amplify noise. The difference map highlights smooth intensity shifts, with no sharp artifacts.

While bias correction helps standardize input intensities for foundation model training, its effects
vary with modality, anatomy, and pathology. Overcorrection may reduce lesion contrast or introduce
distortions, while undercorrection can leave scanner-specific artifacts. Hence, visual and quantitative
validation is essential, particularly when aggregating multi-source data.

\input{tables/pp_bias}

\subsection{Skull Stripping}

\begin{figure}[h]
\centering
\includegraphics[width=\linewidth]{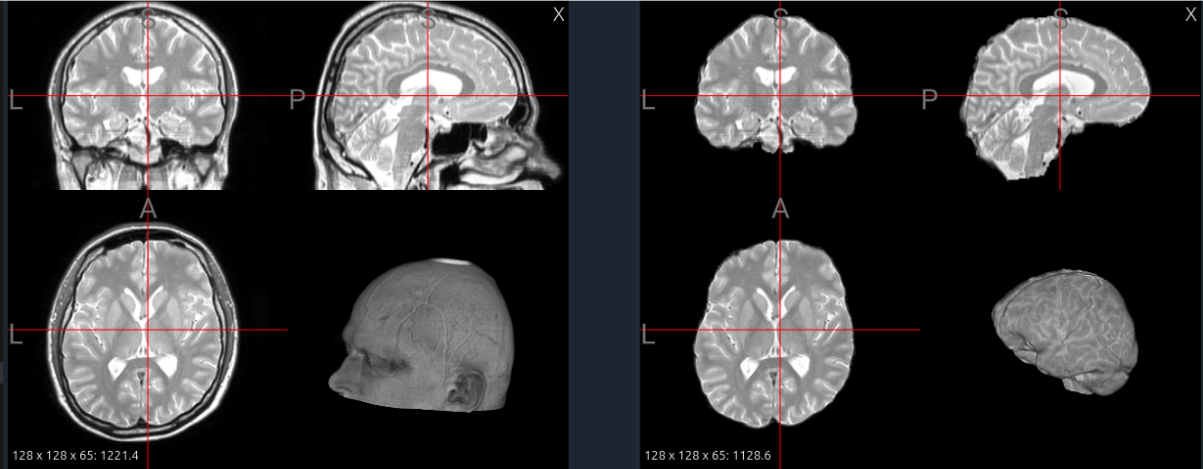}
\caption{Effect of skull stripping on a PD image from the IXI dataset.
Left: original image with visible scalp and skull. Right: stripped image showing improved tissue
contrast and brain boundary definition, but minor over-stripping near cortical edges.}
\label{fig:skull_strip_pdd}
\end{figure}

The primary goal of skull stripping is to remove non-brain tissue, such as the skull, scalp, and dura mater, from the image. This is a critical step as these tissues have high-intensity signals that can interfere with intensity normalization and confuse segmentation algorithms. Common tools include FSL's Brain Extraction Tool (BET) \cite{smith_fast_2002}, AFNI's 3dSkullStrip \cite{cox_afni_1996}, and more recently, deep learning-based methods like HD-BET \cite{isensee_automated_2019}, which often provide more accurate results. While most datasets in our analysis are provided pre-stripped (e.g., BraTS, ISLES22), the specific algorithm used often varies or is not documented, leading to subtle differences in the final brain mask. Figure~\ref{fig:skull_strip_pdd} illustrates the effect of skull stripping on a PD image from the IXI dataset, where non-brain tissues such as the scalp and skull are successfully removed, leaving only the intracranial structures for further analysis.

From a foundation model standpoint, skull stripping can influence both pretraining and downstream transfer. When training models across multiple datasets, consistent skull stripping helps reduce non-biological variability and ensures that the model focuses on relevant brain structures. However, inconsistency across datasets—where some scans are stripped and others are not—can lead to feature-space fragmentation, causing the model to learn dataset-specific biases rather than generalizable brain representations. Therefore, strict harmonization of preprocessing pipelines, including identical skull stripping tools, thresholds, and quality-control procedures, is essential.

Moreover, the choice to strip or retain the skull should align with the model’s target scope. For models designed to capture brain-centric features—such as lesion segmentation, cortical parcellation, or morphometric analysis—skull stripping is generally beneficial, as it directs attention to intracranial tissues. Conversely, for models intended to generalize across multi-modal or multi-organ contexts (e.g., MRI–CT alignment, PET fusion, or structural-to-functional transfer), removing the skull can limit cross-modality correspondence and reduce anatomical completeness. A practical strategy for large-scale foundation model pretraining is to include both stripped and unstripped variants of each scan and use metadata tags or preprocessing embeddings to inform the model about their origin. This dual representation encourages robustness to preprocessing differences and enables the model to learn invariance to skull presence—an increasingly important capability for generalizable medical foundation models.

\subsection{Spatial Registration to MNI152}

Spatial registration aims to align MRI volumes into a common anatomical space, reducing spatial
variability across datasets. Using a modality-aware ANTs pipeline with rigid–affine–SyN transformations,
we aligned representative scans to the MNI152 template. This process standardizes brain geometry but
also exposes how registration can reshape anatomical statistics in subtle, dataset-specific ways.

\begin{figure}[h]
\centering
\includegraphics[width=\linewidth]{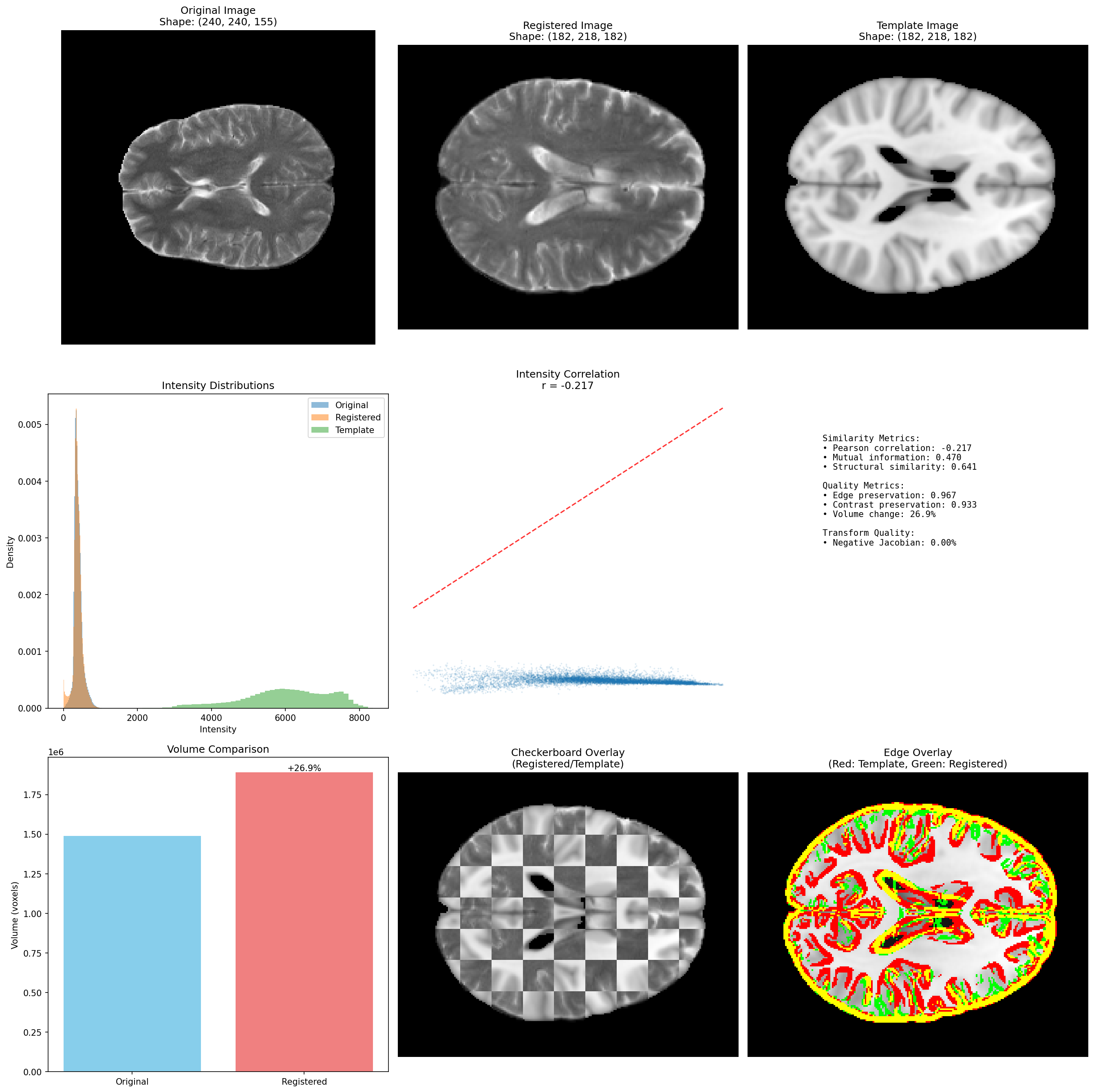}
\caption{Effect of registration on a T2-weighted image from the BraTS-MEN dataset.
Top: original, registered, and MNI152 template images. Bottom: histogram alignment,
intensity correlation ($r=-0.217$), and overlay comparisons showing a 26.9\% volume increase.}
\label{fig:registration_bratsmen}
\end{figure}

Figure~\ref{fig:registration_bratsmen} shows the registration effect for a T2-weighted image from
BraTS-MEN. The aligned scan closely matches the MNI template, and quantitative metrics confirm high
structural similarity (mutual information = 0.974, structural similarity = 0.641). However, resampling
expanded the image volume by 26.9\%, and local correlation ($r = -0.217$) indicates that voxel intensity
relationships were partially altered. Overlay maps and checkerboard comparisons highlight that most
deviations occur near lesion borders and ventricles—regions where pathology or intensity nonuniformity
interact poorly with the template deformation.

These findings reveal an essential trade-off. Registration improves spatial consistency across datasets,
supporting template-based feature extraction and patch sampling. Yet, excessive geometric forcing can
distort pathological anatomy and attenuate lesion contrast, especially in heterogeneous clinical data.
For foundation model pretraining, this suggests that full MNI normalization may be beneficial only for
structural harmonization, while native-space training augmented with local spatial perturbations could
better preserve disease-specific variability and improve cross-domain generalization.

\subsection{Interpolation of Thin-Slice Volumes}

Several clinical datasets, such as MS-60, contain scans with limited z-axis coverage or thick slices,
producing anisotropic volumes that hinder 3D convolutional learning. To mitigate this, we applied an
automated interpolation procedure that increases through-plane resolution while maintaining anatomical
scale. This step is not simply geometric resampling—it directly determines how well small, low-contrast
lesions are represented in 3D feature space.

\begin{figure}[h]
\centering
\includegraphics[width=\linewidth]{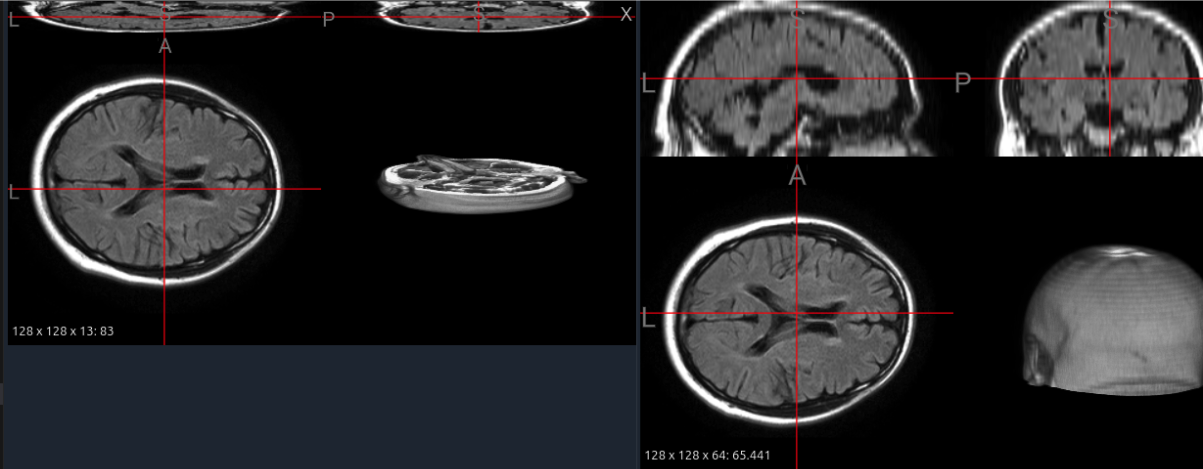}
\caption{Interpolation of a FLAIR image from the MS-60 dataset.
Left: original thin-slice volume (13 slices) showing discontinuities.
Right: interpolated volume (64 slices) with improved z-axis continuity and preserved anatomy.}
\label{fig:interpolation_ms60}
\end{figure}

Figure~\ref{fig:interpolation_ms60} illustrates a FLAIR image from the MS-60 dataset before and after
interpolation. The original scan (13 slices) shows severe discontinuities and collapsed tissue boundaries,
whereas the interpolated version (64 slices) restores smoother cortical contours and continuous sulcal
structures without distorting global shape. Quantitatively, the effective slice thickness decreased by
approximately 4.8$\times$, enabling isotropic patch extraction for pretraining and consistent input
dimensions across datasets.

From a foundation model perspective, interpolation functions as a structural equalizer: it harmonizes
volumetric resolution across sources, improving patch uniformity and kernel receptive fields. However,
it also generates synthetic voxels that may obscure very small hyperintensities or produce interpolation
artifacts along lesion edges. Thus, interpolation should be applied selectively—preferably on
high-anisotropy datasets or in conjunction with uncertainty-aware augmentations—to balance geometric
consistency and lesion fidelity.

%% file: tables/pp_intensity.tex
\begin{table}[h]
\centering
\caption{Voxel-level intensity statistics before and after z-score normalization.}
\label{tab:intensity_stats}
\begin{tabular}{l r r}
\toprule
\textbf{Statistic} & \textbf{Original Image} & \textbf{Normalized Image} \\
\midrule
Minimum & 0.854 & 0 \\
Maximum & 1259.812 & 8.40 \\
Mean & 258.018 & $\sim$0.00 \\
Std & 95.000 & $\sim$1.00 \\
\bottomrule
\end{tabular}
\end{table}

%% file: tables/pp_bias.tex
\begin{table}[h]
\centering
\caption{Quantitative effects of bias field correction on T2-weighted images from BraTS-SSA.}
\label{tab:bfc_stats}
\begin{tabular}{l r r}
\toprule
\textbf{Metric (BraTS-SSA)} & \textbf{Before} & \textbf{After} \\
\midrule
Coefficient of Variation & 0.207 & 0.163 \\
Signal-to-Noise Ratio (SNR) & 6.87 & 6.50 \\
\bottomrule
\end{tabular}
\end{table}

%% file: sections/covariate_shift.tex
\section{Residual Covariate Shift After Preprocessing}

Despite standardized preprocessing, inherent heterogeneity in MRI data from diverse sources introduces residual covariate shift, significantly impeding the generalizability of deep learning models. This phenomenon manifests as subtle, non-linear variations within images, encompassing scanner-specific noise patterns, intensity distortions, and residual artifacts that standard harmonization techniques fail to fully mitigate. To empirically demonstrate this, we utilized T1-weighted MRI scans of healthy subjects from two public datasets: NFBS (125 images) and a subset of IXI (54 images). All data underwent a uniform preprocessing pipeline: skull stripping, N4 bias field correction, MNI152 template registration, and intensity normalization. For computational efficiency and to effectively capture domain differences, a single central axial slice was extracted from each image. Features (1024-dimensional) were subsequently derived from the penultimate layer of an ImageNet-pretrained DenseNet121, without fine-tuning, to assess raw feature transferability.Quantitative assessment of the domain shift between these datasets is summarized in Table \ref{table:covariate_shift}.

Despite a high cosine similarity, indicative of similar vector directions, a substantial Euclidean distance and average Wasserstein distance highlight significant shifts in the magnitude and distribution of features. Statistical analysis further confirmed this divergence: 83.89\% of all features exhibited statistically significant differences ($p < 4.88\times10^{-5}$) after Bonferroni correction). These findings conclusively demonstrate that standard preprocessing is insufficient for complete MRI data harmonization. The persistent residual covariate shift in the learned feature space critically impairs model robustness and transferability across unseen domains. Therefore, developing and implementing explicit domain adaptation strategies—such as disentangled representation learning, meta-learning for domain generalization, and robust uncertainty estimation—is paramount for building truly generalizable and clinically reliable models. This is particularly crucial for the advancement of foundation models in high-stakes medical imaging applications.

\input{tables/covariate_shift}

%% file: tables/covariate_shift.tex
\begin{table}[h]
\centering
\caption{Quantitative assessment of residual covariate shift in learned feature space: NFBS vs. IXI Datasets}
\label{table:covariate_shift}
\begin{tabular}{l r}
\toprule
\textbf{Metric} & \textbf{Value} \\
\midrule
Cosine Similarity (mean vectors) & 0.960907 \\
Euclidean Distance (mean vectors) & 6.854918 \\
Average 1D Wasserstein Distance & 0.141903 \\
Significant Features ($p < 0.05$) & 947 / 1024 \\
Bonferroni-Significant Features ($p < 4.88 \times 10^{-5}$) & 859 / 1024 \\
\bottomrule
\end{tabular}
\end{table}

%% file: sections/conclusion.tex
\section{Conclusion \& Discussion}

In this study, we analyzed \totaldatasets{} publicly available brain MRI datasets to understand how their characteristics differ across modalities, voxel geometry, and intensity distributions. We found that public datasets cover a broad range of neurological and psychiatric conditions but vary widely in scale, modality composition, and acquisition settings. Structural MRI sequences dominate the landscape, while advanced modalities such as diffusion and functional MRI are much less represented. This diversity provides valuable opportunities for comprehensive modeling but also poses challenges for developing foundation models that must generalize across many domains. To assess how preprocessing affects data harmonization, we applied a standardized pipeline including bias-field correction, intensity normalization, skull stripping, and spatial registration. These steps increased internal consistency within datasets but did not fully remove differences between them. Residual variability was evident in the feature space, indicating that standard preprocessing alone cannot ensure complete harmonization. This suggests that preprocessing-aware model design and domain adaptation techniques are needed to reduce inter-dataset shifts more effectively.

Several limitations should be acknowledged. Our study did not assess annotation quality or consistency, which strongly influences the suitability of datasets for supervised learning. We also did not perform model benchmarking, which would show how dataset variability translates to performance differences. Future work will address these aspects by analyzing annotation quality, increasing the number of samples for preprocessing evaluation, and including model benchmarking. Together, these efforts will build a stronger foundation for developing harmonized, reliable, and generalizable brain MRI foundation models.

%% file: sections/acknowledgement.tex
\section*{Author Declaration}
This manuscript was prepared and refined with the assistance of ChatGPT (GPT-5, OpenAI, 2025) for language enhancement and clarity. 

\section*{Acknowledgement}
This work was supported by a grant for research centers, provided by the Ministry of Economic Development of the Russian Federation in accordance with the subsidy agreement with the Novosibirsk State University dated April 17, 2025 No. 139-15-2025-006: IGK 000000C313925P3S0002

%% file: sections/appendix.tex
\appendix
\renewcommand{\thetable}{A\arabic{table}}
\renewcommand{\thefigure}{A\arabic{figure}}
\setcounter{table}{0}
\setcounter{figure}{0}

\section*{Appendix}

\renewcommand{\arraystretch}{1.2}

\begin{table}[h]
\centering
\caption{Standardization of imaging modalities across datasets.}
\label{tab:modality_standardization}
\begin{tabular}{p{4cm} p{8cm}}
\hline
\textbf{Standardized Label} & \textbf{Original Variants Grouped} \\
\hline
T1 & Anatomical T1-weighted scans \\
T1c & Contrast-enhanced T1-weighted scans \\
DWI/DTI & DWI, DTI, ADC, TRACE \\
fMRI/rs-fMRI & BOLD, resting-state fMRI, task-fMRI \\
Others & SWI, ASL, PET, CT, MRA, MEG \\
\hline
\end{tabular}
\end{table}

\begin{table}[h]
\centering
\caption{Standardization of cohort labels across datasets.}
\label{tab:cohort_standardization}
\begin{tabular}{p{4cm} p{8cm}}
\hline
\textbf{Standardized Label} & \textbf{Original Variants Grouped} \\
\hline
Healthy & Healthy \\
Stroke & Stroke \\
Multiple Sclerosis & Multiple Sclerosis \\
Brain Tumor & Glioma, Glioblastoma, Meningioma, Metastasis \\
Neurodegenerative & Alzheimer’s Disease, Parkinson’s Disease \\
Psychiatric Disorder & Schizophrenia, Bipolar Disorder, Major Depressive Disorder (MDD) \\
\hline
\end{tabular}
\end{table}

\begin{figure}[h]
\centering
\includegraphics[width=\linewidth]{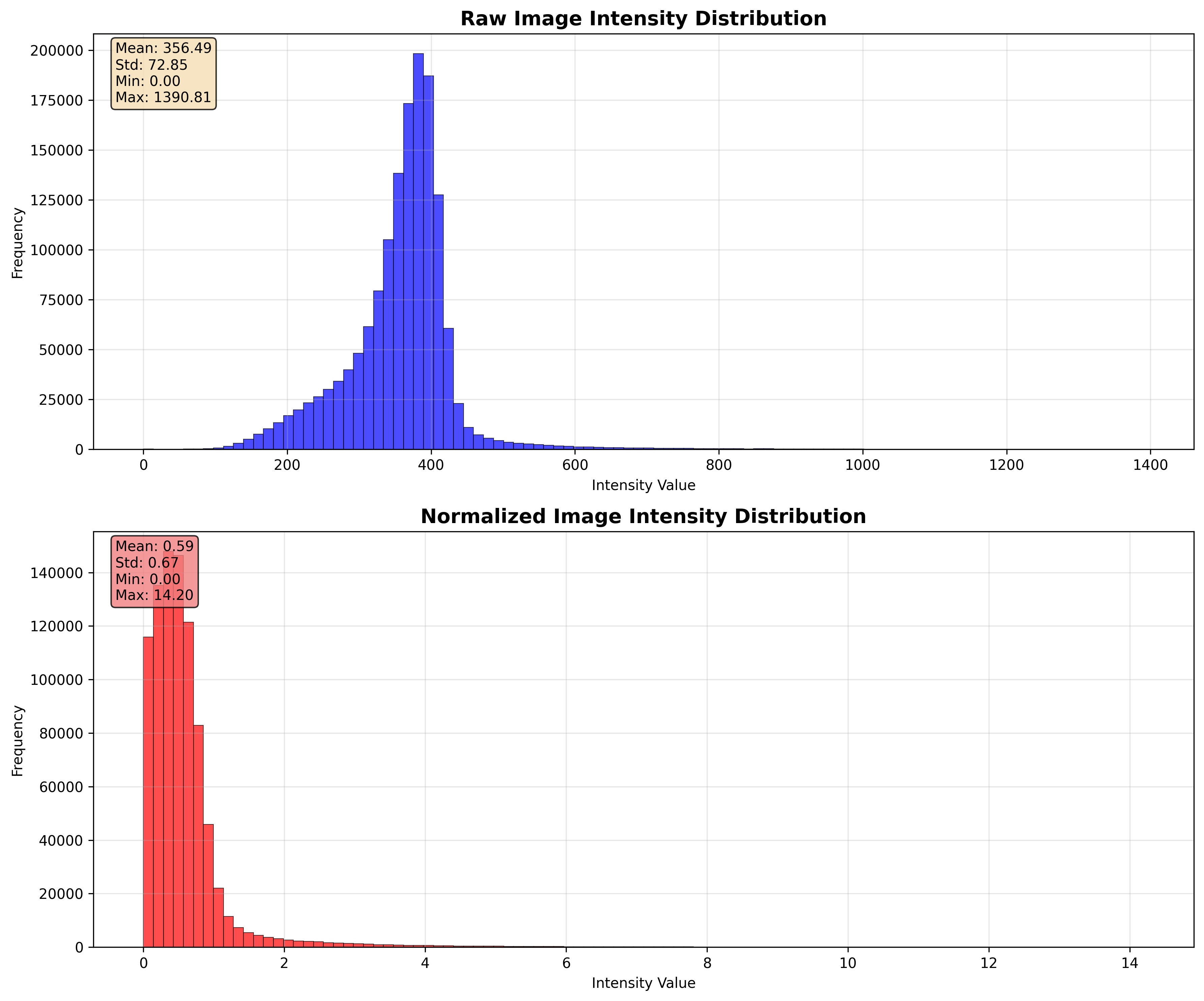}
\caption{Z-score intensity normalization effect, T1C from BraTS-MEN dataset.}
\label{fig:zscore_dist}
\end{figure}

\begin{figure}[h]
\centering
\includegraphics[width=\linewidth]{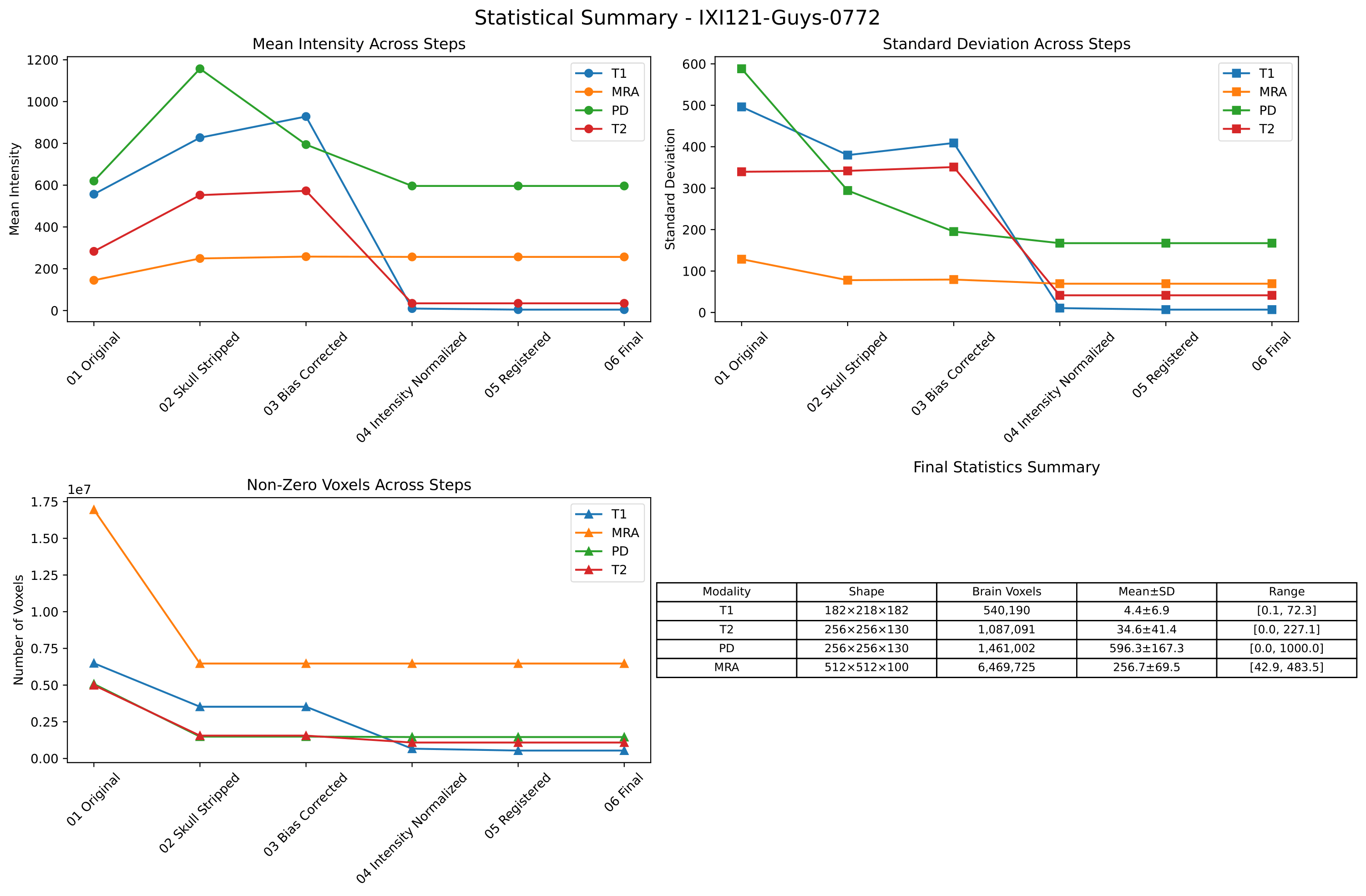}
\caption{Statistical summary of intensity distributions across steps for IXI dataset. Preprocessing reduces variability, especially in outlier values.}
\label{fig:ixi_stats}
\end{figure}